\documentclass[usletter, conference]{style/IEEEtran}
\usepackage{times}

\usepackage[numbers]{natbib}
\usepackage{multicol}
\usepackage[bookmarks=true]{hyperref}
\usepackage{multirow}
\usepackage[pdftex]{graphicx}
\usepackage{graphics,graphicx} 
\usepackage{epsfig} 
\usepackage{mathptmx} 
\usepackage{times} 
\usepackage{amsmath, bm} 
\usepackage{amssymb}  
\usepackage{algorithm}
\usepackage{algorithmic}
\usepackage[dvipsnames]{xcolor}
\usepackage{url} 
\usepackage{ifpdf}
\usepackage{amsfonts}
\usepackage{mathtools}
\usepackage{verbatim}
\usepackage{caption}
\usepackage{subcaption}

\begin{document}
\title{\LARGE \bf The Actuation-consistent Wrench Polytope (AWP) and the Feasible Wrench Polytope (FWP)}
\author{$^{1}$Romeo Orsolino, $^{1}$Michele Focchi, $^{1}$Carlos Mastalli,\\ $^{2}$Hongkai Dai, $^{1}$Darwin G. Caldwell and $^{1}$Claudio Semini\\
$^{1}$Department of Advanced Robotics, Istituto Italiano di Tecnologia (IIT), Genova, Italy\\
$^{2}$Toyota Research Institute (TRI), Los Altos, USA
}

\maketitle
\thispagestyle{empty}
\pagestyle{empty}
\vspace{-.5cm}
\section{Motivation}
The motivation of our current research is to devise motion planners for legged locomotion that are able to exploit the robot's actuation capabilities. This means, when possible, to minimize joint torques or to propel as much as admissible when required. For this reason we define two new 6-dimensional bounded polytopes that we name Actuation-consistent Wrench Polytope (AWP) and Feasible Wrench Polytope (FWP).
\section{Introduction}
We define the former polytope (the AWP) as the set of all wrenches that a robot can generate on its own center of mass (CoM). This considers the contact forces that the robot can generate given its current configuration and actuation capabilities. Unlike the Contact Wrench Cone (CWC) \cite{Hirukawa2006, Dai2016}, the AWP is a bounded polytope in $\in {\rm I\!R}^6$.\\
The intersection of the AWP with the CWC results in the second convex bounded polytope $\in {\rm I\!R}^6$ that we define as Feasible Wrench Polytope (FWP). This contains all the feasible wrenches that can be realized on the robot's CoM, both given the \textit{internal} properties of the robot (its posture and its actuation limits) and the \textit{external} properties coming from the contact with the environment (unilaterality, normal direction, friction cone coefficient).
\begin{figure}[h]
\includegraphics[scale=.15]{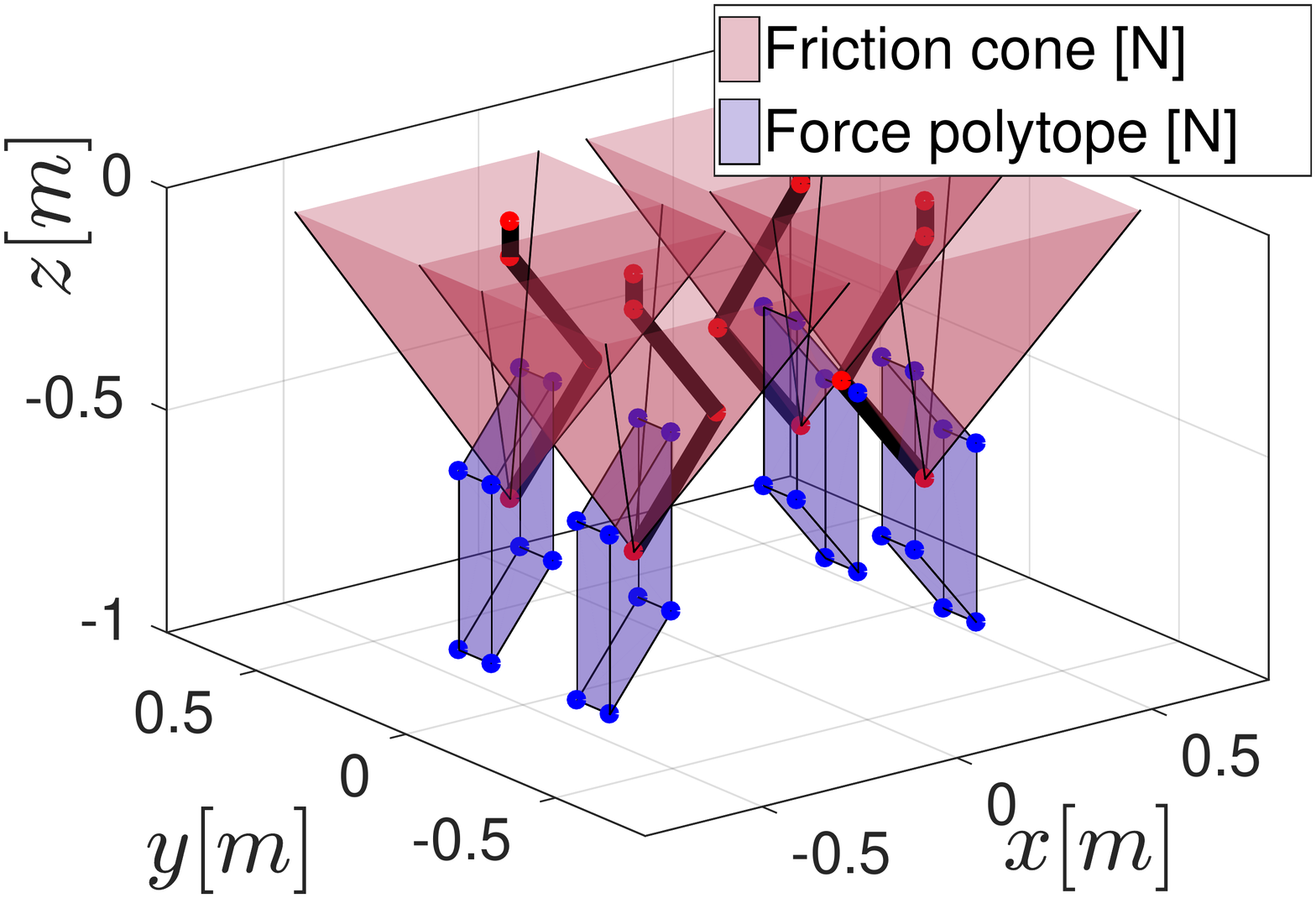}
\includegraphics[scale=.14]{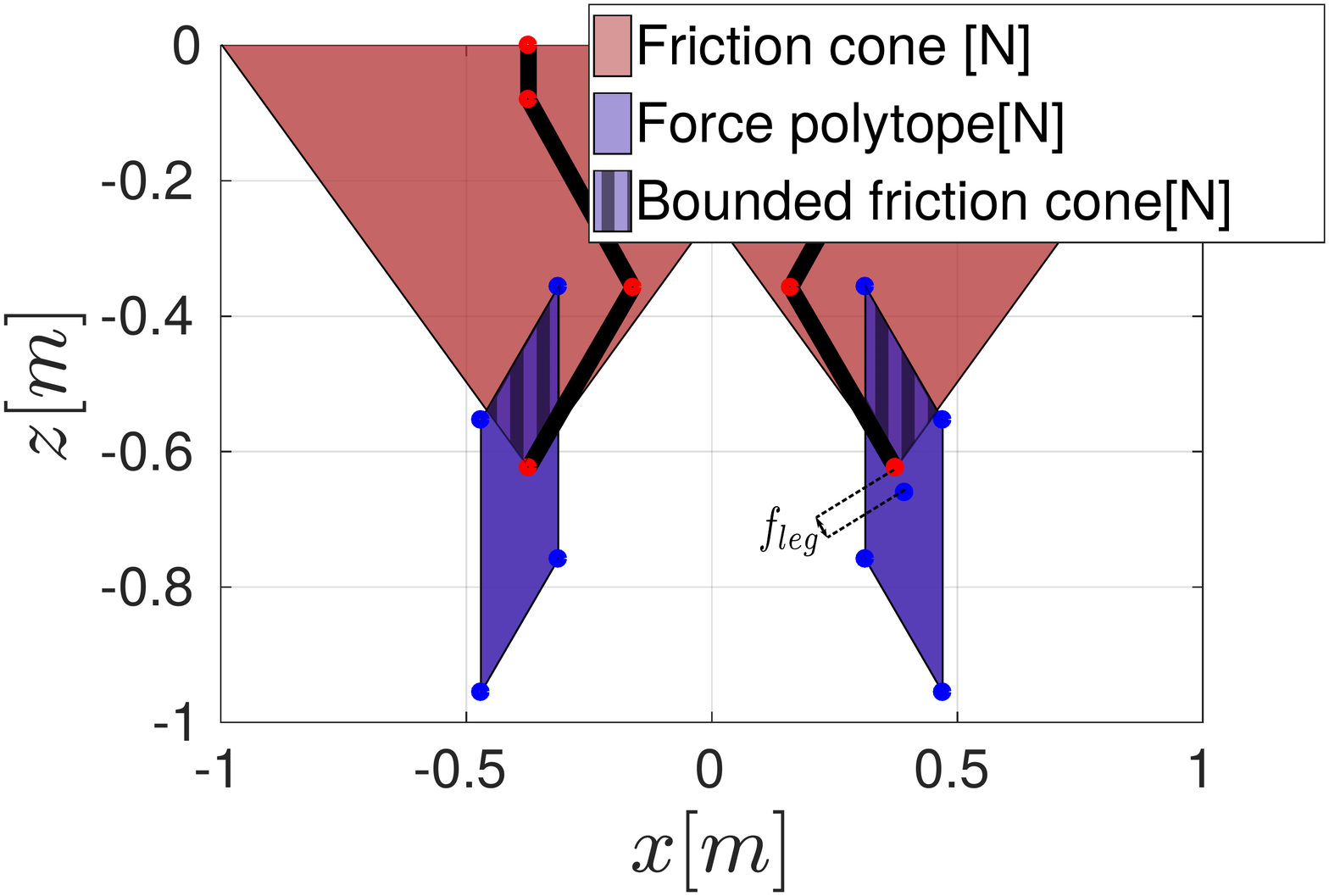}
\caption{3D force polytopes (blue) and linear friction cones (pink). The intersection of these two sets forms 3D \textit{bounded friction polytopes} that contain all the feasible contact forces that can be generated at each foot.}
\label{fig:fws_2}
\end{figure}
\vspace{-.5cm}
\begin{figure}[h!]
\includegraphics[scale=.15]{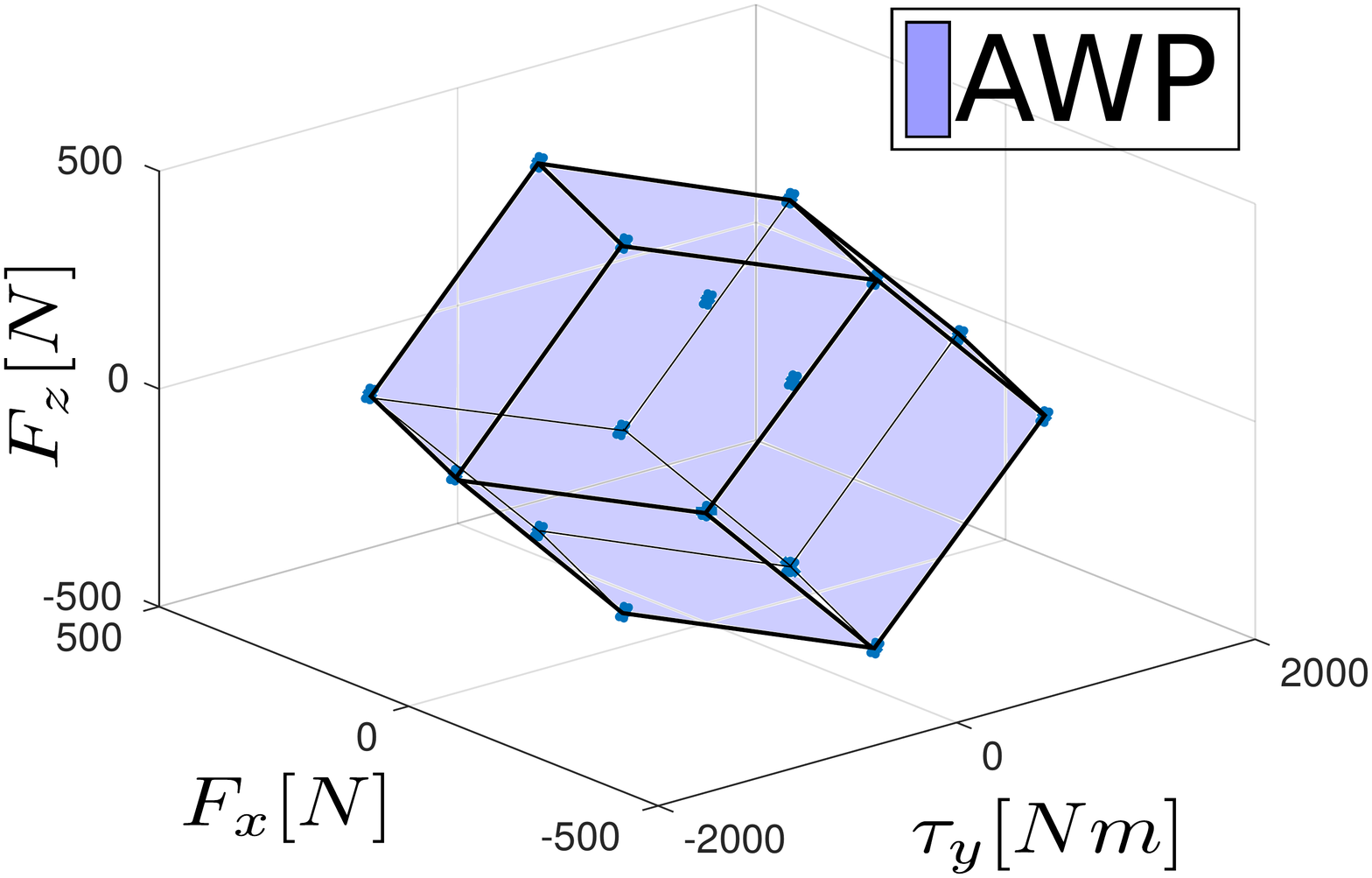}
\includegraphics[scale=.14]{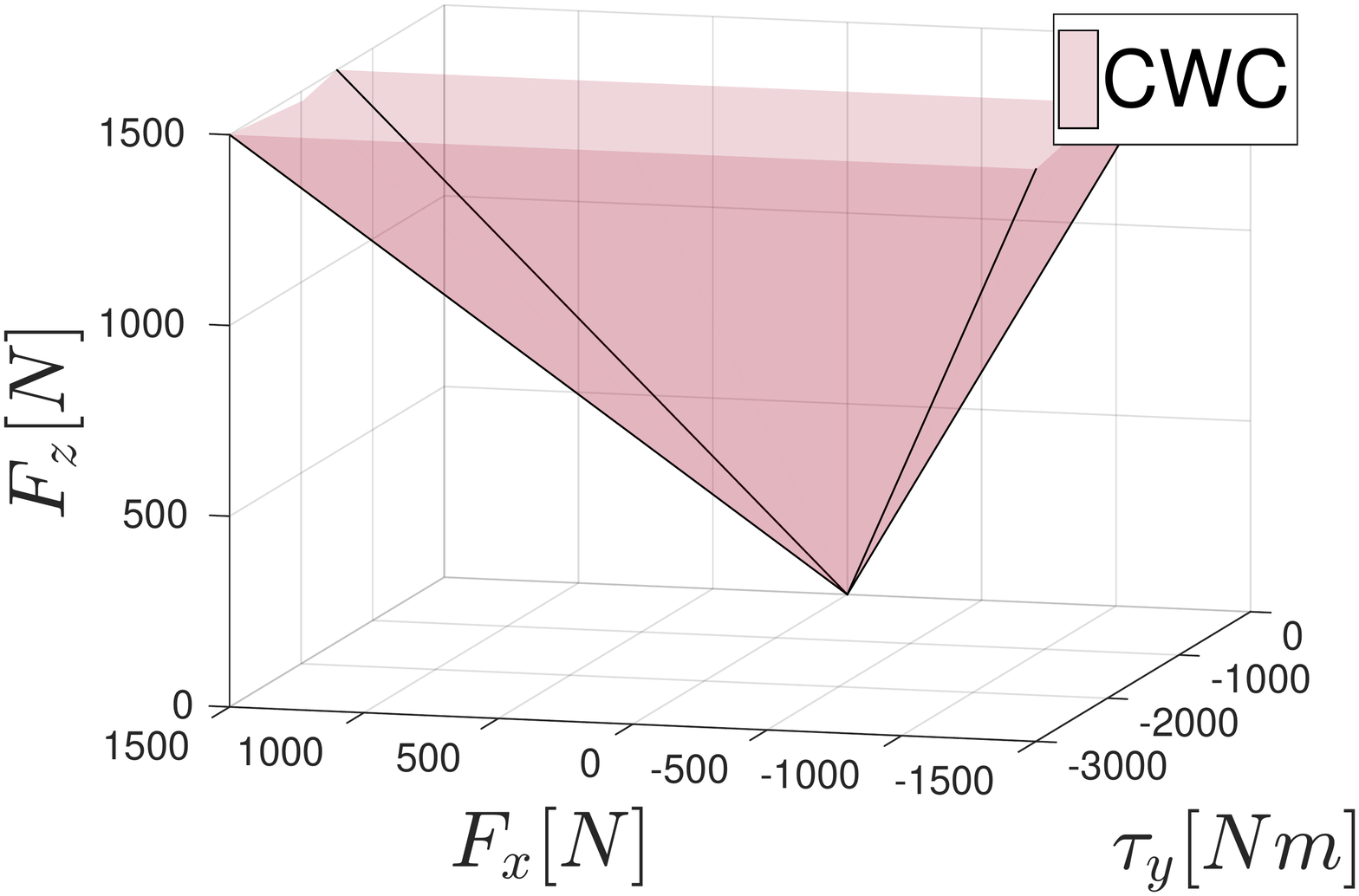}
\caption{Actuation Wrench Polytope (AWP) (left) and Contact Wrench Cone (CWC) (right) in a the case of a planar model. The FWP is the intersection of the AWP and the CWC.}
\label{fig:tls_2}
\end{figure}
\section{The Actuation Wrench Polytope (AWP)}
The AWP is a bounded convex polytope in ${\rm I\!R}^6$ that contains all the wrenches that a robot can apply given its torque limits and its current configuration.\\
Let us consider a tree-structured robot with $n_b$ branches: the first step for computing the AWP is to compute the upper and lower bounds of the contact forces $\mathbf{f}^{max} \in {\rm I\!R}^{3}$ that each end effector of robot can realize on the environment. This can be performed using the following relationship:
\vspace{-.1cm}
\begin{equation}\label{eq:1}
\mathbf{f}^{\max}_i = \mathbf{J}_i^{T^{\#}} \big(\mathbf{M}_{bi}^T
\mathbf{\ddot{q}}_b + \mathbf{M}_i \mathbf{\ddot{q}}_i + \mathbf{c}_i + \mathbf{g}_i -
\mathbf{\tau}_i^{\max}\big),
\end{equation}
where $\mathbf{\tau}^{max}_i \in {\rm I\!R}^{n}$ is a vector containing the torques generated by the $n$ actuated joints $\mathbf{q}_i$ of the robotic limb and $\mathbf{q}_b$ represents the degrees of freedom of the base. $\mathbf{g}_i$ accounts for the gravity acting on the limbs and $\mathbf{c}_i$ is the centrifugal/coriolis term. $\mathbf{J}^{\#}(q)$ is the Moore-Penrose pseudoinverse of the trasposed jacobian matrix of the leg: $\mathbf{J}^{\#} = (\mathbf{J}(\mathbf{q}) \mathbf{J}(\mathbf{q})^T)^{-1} \mathbf{J}(\mathbf{q})$.\\
Since each element of $\mathbf{\tau}^{max}$ can take on two values (either the maximum or the minimum torque) then there are $2^n$ combinations that give origin to $2^n$ different values of $\mathbf{f}^{max}_i$ with $i = 1,2,\dots, n$. Each $\mathbf{f}^{max}_i \in {\rm I\!R}^3$ is a vertex of a 3D force polytope \cite{Chiacchio1997} that represents the set of all possible contact forces that can be generated by the limb in that configuration. Fig. \ref{fig:fws_2} represents such 3D force polytopes composed by the torque limits (blue) and by the linear friction cones (pink) for a quadruped robot.\\ 
The second step for the computation of the AWP is to add three more dimensions to each of these 3D points $\mathbf{f}^{max}_i$:
\begin{equation}
\mathbf{w}_i = \left(
\begin{bmatrix}
    \mathbf{f}^{max}_i \\
    \mathbf{p}_k \times \mathbf{f}^{max}_i 
\end{bmatrix} \right)  \in {\rm I\!R}^6 \hspace{.5cm} with: i = 1,2,\dots, 2^n
\end{equation}
with $\mathbf{p}_k \in {\rm I\!R}^3$ being the position of the $k^{th}$ end-effector. \\
The new dimensions add the corresponding torque that the robot can generate on the its own CoM. \\ 
In this way it is possible to obtain a 6D bounded polytope for each of the $n_b$ limbs of the tree-structured robot: $\mathcal{W}_{k} = ConvHull(\mathbf{w}_1^k,\mathbf{w}_2^k,\dots,\mathbf{w}_{2^n}^k)$ with $k = 1,2,\dots n_b$.\\
The final step to compute the $AWP$ is then to perform the sum of the 6D wrench polytopes $\mathcal{W}_k$ of each limb: 
\begin{equation}
AWP = \mathcal{W}_1 \bigoplus \mathcal{W}_2 \bigoplus \dots \mathcal{W}_{n_b}
\end{equation}
Fig. \ref{fig:tls_2} shows the AWP and the CWC in the case of a planar model, whose full wrench space is described by the variables $F_x, F_z$ and $\tau_y$ and can thus be represented in 3D.
\section{The Feasible Wrench Polytope (FWP)}
The goal of the FWP is to remove from the AWP all the contact forces that are not feasible because of the constraints given by the environment. These features are the fact that the leg can only push and not pull (unilaterality) and the fact that, to avoid slippage, the contact force should stay within the friction cone. The computation of these constrains requires the knowledge of the surface normal and of the friction coefficient at each contact point. \\
The FWP can be obtained by intersecting the AWP with the CWC. This is computationally expensive because of the high cardinality of the two sets (high number of half-spaces). For this reason we analyze possible approaches (e.g. exploiting the $V$-description, the $H$-description or the double-description \cite{Caron2015}) that can allow to use AWP and the FWP for \textit{online} motion planning.
\section{Conclusion}
The purpose of the AWP and of the FWP is to have a description as precise as possible of the admissible states that a legged robot can reach. The FWP can be used to devise controllers that exploit the all actuation range of a robot in order to, by instance, reject external disturbances on terrains of arbitrary roughness.\\
The FWP can also be exploited to design motion planners that minimize the actuation torques of each joint while still using a simplified centroidal model of the robot. \\
Besides that, the concept of AWP and FWP that has been described so far for the specific case of a legged robots, can be extended to other types of platforms that share a similar complexity of interaction with the environment (such as robotic manipulators for grasping).\\
Our current research focuses on assessing the performance of \textit{online} motion planners that only use the vertex-description of the 6D sets \cite{Delos2015}.
\bibliographystyle{unsrt} 
\bibliography{references/refs_abstract}

\begin{thebibliography}{1}

\bibitem{Hirukawa2006}
H.~Hirukawa, K.~Kaneko, S.~Hattori, F.~Kanehiro, K.~Harada, K.~Fujiwara,
  S.~Kajita, and M.~Morisawa.
\newblock {A Universal Stability Criterion of the Foot Contact of Legged Robots
  - Adios ZMP}.
\newblock {\em IEEE ICRA}, 2006.

\bibitem{Dai2016}
H.~Dai.
\newblock {Robust Multi-Contact Dynamical Motion Planning using Contact Wrench
  Set}.
\newblock {\em PhD thesis}, 2016.

\bibitem{Chiacchio1997}
P.~Chiacchio, Y.~Bouffard-Vercelli, and F.~Pierrot.
\newblock {Force Polytope and Force Ellipsoid for Redundant Manipulators}.
\newblock {\em Journal of Robotic Systems}, 14(8):613--620, 1997.

\bibitem{Caron2015}
S.~Caron, Q.-C. Pham, and Y.~Nakamura.
\newblock {Leveraging Cone Double Description for Multi-contact Stability of
  Humanoids with Applications to Statics and Dynamics}.
\newblock {\em Robotics: Science and System (RSS)}, 2015.

\bibitem{Delos2015}
V.~Delos and D~Teissandier.
\newblock Minkowski sum of polytopes defined by their vertices.
\newblock {\em Journal of Applied Mathematics and Physics (JAMP)}, 2015.

\end{thebibliography}
\end{document}